# The Empowerment of Science of Science by Large Language Models: New Tools and Methods

Guoqiang Liang1, Jingqian Gong1, Mengxuan Li2, Gege Lin1,*, Shuo Zhang1,*

(1.College of Economics and Management, Beijing University of Technology, Beijing, 100124, China. 2.College of Economics and Management, Langfang Normal University, Langfang, Hebei 065000, China)

**ORCID of the authors**
Guoqiang Liang: 0000-0002-9669-4048
Shuo Zhang: 0000-0001-7251-2943

**Corresponding author:**
Gege Lin, College of Economics and Management, Beijing University of Technology, Beijing, 100124, China. phone: 18335915377, E-mail: lingege@bjut.edu.cn
Shuo Zhang, College of Economics and Management, Beijing University of Technology, Beijing, 100124, China. phone: 17710981334, E-mail:zhangsshuozs@bjut.edu.cn

**Acknowledgements:** Not applicable

**Statements and Declarations**
**Declaration of conflicting interest:** The authors declared no potential conflicts of interest with respect to the research, authorship, or publication of this article.
**Funding statement:** The authors disclosed receipt of the following financial support for the research, authorship, and/or publication of this article: This work was support by the National Natural Science Foundation of China (Project No. 72204014 & 72304023), and the Beijing Natural Science Foundation, China (Project No. 9232002).
**Ethical approval and informed consent statements**:This article does not contain any studies with human or animal participants.
**Data availability statement**: The code and data are available on the following sites:
https://github.com/Gqiang-Liang/Simple-demo-for-NRE/tree/main
https://chatglm.cn/main/gdetail/6632ecfeace21f9ff21cf4c0?lang=zh

**Abstract** Large Language Models (LLMs) have exhibited exceptional capabilities in natural language understanding and generation, image recognition, and multimodal tasks, charting a course towards Artificial General Intelligence and emerging as a central issue in the global technological race. This manuscript conducts a comprehensive review of the core technologies that support LLMs from a user's standpoint, including prompt engineering, knowledge-enhanced retrieval-augmented generation (RAG), fine-tuning, pre-training, and tool learning. Additionally, it traces the historical development of Science of Science (SciSci) and presents a forward-looking perspective on the potential applications of LLMs within the scientometric domain. Furthermore, it discusses the prospect of an AI agent-based model for scientific evaluation, and presents new research fronts detection and knowledge graph building methods with LLMs.

The rapid development of large-scale pre-trained models, often referred to as "large language models (LLMs)", has driven a transformation in the field of artificial intelligence. These models have demonstrated superior performance in natural language understanding and generation, image recognition, and even multimodal tasks, paving the way for Artificial General Intelligence (AGI) and becoming a focal point in global technological competition. As of 2023, the United States leads with 61 LLMs, significantly outpacing the European Union's 21 and China's 15 LLMs, making it a top nation in artificial intelligence. These models have been widely applied in various sectors, including healthcare, finance, education, law, and mathematics [1].

SciSci is an interdisciplinary field that conducts quantitative research on science itself. Its scope includes scientists, academic literature, scholarly journals, and science policies. The methodologies used in SciSci have been evolving from traditional citation analysis, word frequency analysis, and statistical analysis to incorporate computer science and artificial intelligence. For instance, scholars have begun to use dynamic topic models and word2vec for analyzing semantic-enhanced topic evolution [2]. BERT models [3], Graph Convolutional Networks (GCNs) [4], and knowledge graphs [5] are now employed for citation recommendation, keyword identification, identifying emerging technological themes [6], and exploring the research characteristics of Nobel laureates [7].

This study systematically reviews the underlying technical architecture, common concepts, capabilities of large models, as well as the key technologies underpinning LLMs from the users' perspective. Then, the study illustrates the evolution of SciSci from past to present. Finally, provide the prospective applications of LLMs within the field of SciSci.

**1 Introduction to LLMs**

LLMs, also known as "foundation models," are deep neural network models with a vast number of parameters and complex computational structures. They are characterized by their scalability (large parameter volume), emergent properties (the ability to develop new capabilities unexpectedly), and universality (not limited to specific problems or domains). These models are capable of driving multiple use cases and applications, as well as resolving various tasks, making them "milestones" in the fields of natural language processing (NLP) and artificial intelligence.

Similar to the human brain, LLMs, due to their enormous number of parameters and deep neural network architecture, can learn and understand a broader range of features and patterns. This enables them to demonstrate remarkable capabilities in natural language understanding and generation, reasoning, intent recognition, and the creation of images and videos from text, covering virtually all aspects related to NLP. They also possess general problem-solving abilities and are considered a significant path towards achieving general artificial intelligence [8]. Currently, LLMs have become the infrastructure of the AI field, providing powerful computational, learning, and problem-solving capabilities for addressing a variety of complex issues. Examples include weather forecasting [9], behavioral analysis [10], and drug synergy [11], effectively accomplishing complex modeling and predictive tasks.

The massive data input and Transformer architecture was the main capability source of LLMs. Taking OpenAI's GPT series as an example (Table 1 is an extended version based on reference [12]), in 2018, OpenAI introduced the GPT-1 model, which was based on a 12-layer Transformer architecture and trained on approximately 5GB of data. This model significantly improved computational speed and model capacity compared to long short-term memory (LSTM) models, marking a major advancement in Transformer-based architectures. The following year, OpenAI built upon the GPT-1 to release GPT-2, featuring a 48-layer Transformer architecture and trained on data eight times larger than GPT-1. This allowed the model to better understand semantics and contextual information, demonstrating formidable text generation capabilities.

In 2020, OpenAI released the GPT-3 model, based on the GPT-2 architecture, with the number of Transformer layers doubled and the amount of pre-training data increased by over a thousand times. GPT-3 enabled user interaction through natural language and was capable of performing most NLP tasks such as automatic question answering, text classification, and machine translation, showcasing astonishing natural language understanding abilities. It wasn't until the emergence of ChatGPT that the academic community realized the disruptive potential of LLMs on traditional paradigms of natural language processing tasks. The introduction of ChatGPT-4 has further propelled multimodal LLMs to the forefront of cutting-edge research today.

Table 1 Pre-trained Data Volume of ChatGPT

| Models | Architecture | Layers | Parameters | Data size |
| --- | --- | --- | --- | --- |
| GPT-1(2018) | Transformer | 12 | 110 milion | 5GB |
| GPT-2(2019) | Transformer | 48 | 1.5 billion | 40GB |
| GPT-3(2020) | Transformer | 96 | 175 billion | 45TB |
| ChatGPT-4(2023) | Transformer | 120 | 1.76 trillion | Not disclosed |

**1.1 Classification of LLMs**

LLMs can be classified into different types based on various criteria. For instance, when categorized by input data type, LLMs can be divided into language models, visual models, and multimodal models. Language models are primarily used for processing text data and understanding natural language, making them a significant category within the field of NLP. These models are characterized by their training on large-scale corpora to learn various grammatical, semantic, and contextual rules of natural language, such as GPT-3, Bard, ERNIE Bot, and ChatGLM, etc.

Visual models are typically used for image processing and analysis, being commonly employed in the field of computer vision (CV). These models are trained on extensive image datasets to perform visual tasks such as image classification, object detection, image segmentation, pose estimation, and facial recognition. Examples of these models include the VIT series (Google), Wenxin UFO, Huawei Pangu CV, and INTERN (SenseTime), etc. As for multimodal models, they combine features of both language and visual models, enabling them to process text, images, and videos simultaneously for a more comprehensive understanding of complex data. Examples of multimodal models include ChatGPT-4, Sora, and Gemma2, etc. Figure 1 illustrates the parameters and classification of notable LLMs.

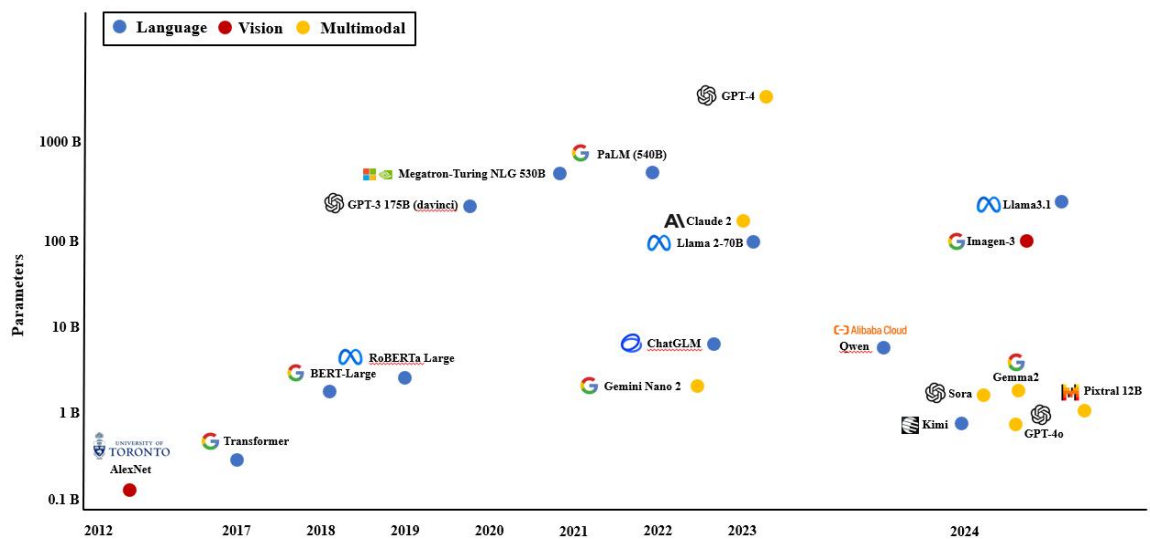

Figure 1. Parameters and classification of notable LLMs

Based on different model architectures, LLMs can be divided into those based on the Transformer architecture and those using the Mixture of Experts (MoE) architecture. LLMs that utilize the Transformer architecture are designed around the Transformer model, introduced by Vaswani et al. in 2017 [13]. This architecture leverages the self-attention mechanism, allowing it to effectively handle long-distance dependencies in sequential data.

The core components of the Transformer model are multi-head self-attention and positional encoding, enabling the model to capture relationships between different positions in the input sequence. Due to its outstanding performance, the Transformer has become the foundational architecture for many large language models, such as BERT and the GPT series.

On the other hand, MoE models are a type of distributed expert system that assigns tasks to multiple "expert" subnetworks [14]. A gating network determines which expert should handle each input sample. This architecture allows models to scale to very large sizes, as increasing the number of experts can enhance the model's capacity and performance without significantly increasing the complexity of any individual expert. MoE models have demonstrated superior scalability and efficiency in handling certain tasks, such as language modeling and image recognition.

Both architectures have their advantages: the Transformer architecture is widely used in NLP tasks due to its efficiency in processing sequential data, while the MoE architecture has garnered attention for its scalability and parallel processing capabilities. Additionally, based on the application domain, LLMs can be categorized into general-purpose models and vertical models; according to the type of auto encoder, they can be further divided into encoder-based models and decoder-based models, among others, which will not be detailed here.

## 1.2 Common Terminology for LLMs

With the development of AI, various concepts such as general-purpose models, vertical models, fine-tuning, tokenization, embedding, and AI agents have emerged [15-18]. These concepts can be easily confused, which is why Figure 2 provides an overview of the relationships between them.

In simple terms, LLMs can be categorized into general-purpose models and vertical models. General-purpose models are pretrained on large public datasets, while vertical models are primarily fine-tuned based on specific domain or industry data using general-purpose models as a foundation [19]. To enhance model performance on specific tasks, the process of further training using labeled data is known as fine-tuning [20, 21]. Fine-tuning or pretraining is predicated on tokenization and embedding, where input data is mapped into a high-dimensional vector space for computation. An AI agent is an intelligent entity based on LLMs, and equipped with planning, memory, and tool-learning capabilities. Figure 2 illustrates the relationships among the main concepts related to LLMs.

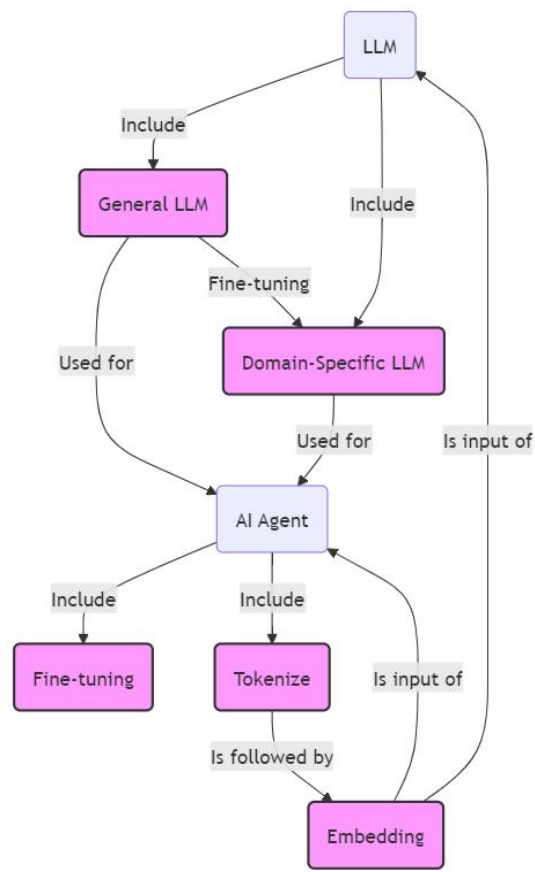

Figure 2. The common terminology and relation of LLMs

In summary, we can view LLMs as a type of neural network-based auto-regressive language model. Essentially, they function as probabilistic language models that learn language patterns from vast amounts of corpus data and output the most likely correct answers based on user input.

**1.3 Workflow of LLMs**

A typical model based on the Transformer architecture generally processes input data in three steps. First, the input data undergoes embedding, which includes both word embedding and position embedding. After the input text is tokenized, each token is transformed into a high-dimensional vector using word embedding techniques. These high-dimensional vectors are then concatenated with position embedding vectors, which capture the position of the tokens in the text.

Second, the concatenated data is passed through multiple Transformer layers. During this process, the self-attention mechanism plays a key role in understanding semantic relationships. We can represent the self-attention mechanism with Equation 1, where Q denotes query, K denotes Key, V denotes Value [22, 23]. Finally, the model predicts the most likely next token in the sequence based on the context and continues generating subsequent tokens through an autoregressive approach, completing the text generation task. In summary, the basic workflow of LLMs and key information about the self-attention mechanism are shown in Figure 3. The detailed process of the self-attention mechanism, as depicted in Figure 3B, can be found in reference [13]

$$\text{Attention}(Q,K,V)=\text{softmax}\left(\frac{QK^T}{\sqrt{d_K}}\right)*V \quad \text{(Equation 1)}$$

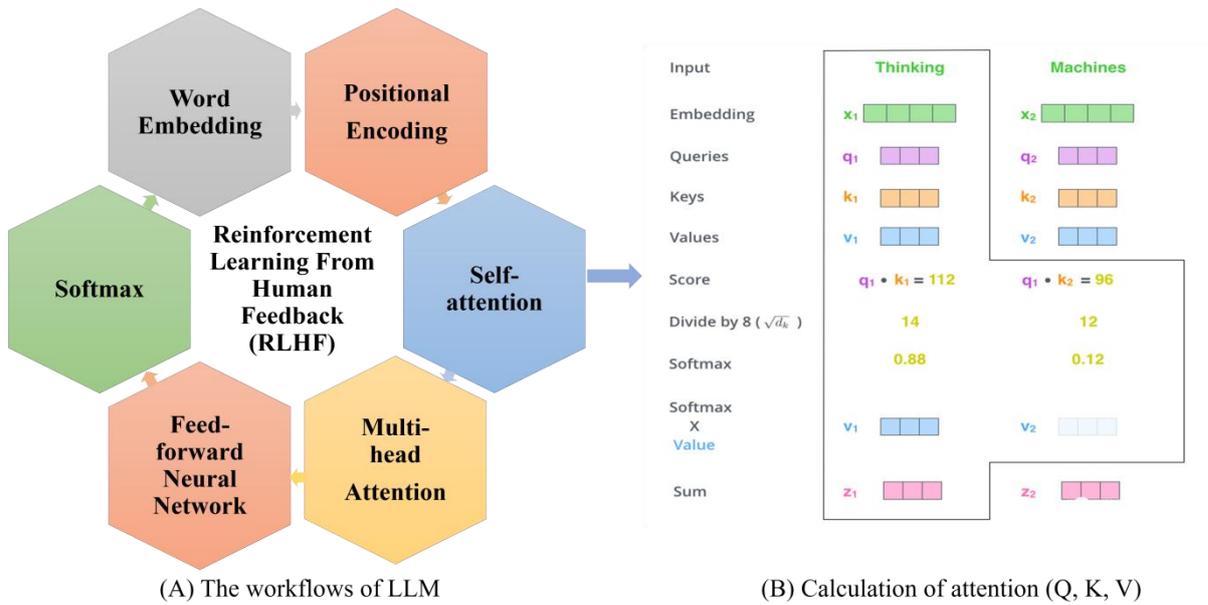

(A) The workflows of LLM    (B) Calculation of attention (Q, K, V)

Figure 3. Workflow of LLMs and analysis of the self-attention mechanism

## 1.4 Key techniques of LLMs

From a user's perspective, there are five key technologies associated with LLMs: prompt engineering, knowledge-enhanced retrieval-augmented generation (RAG), fine-tuning, pre-training, and tool learning, respectively. These five technologies generally increase in complexity.

Prompt engineering involves designing or optimizing the input prompts to guide LLMs in generating outputs that meet the user's expectations, thereby allowing LLMs to better serve user needs [24]. Essentially, a prompt is a text-based input to the LLMs to guide its output. When the input is in the form of speech, the LLMs first convert it into text, which is then used as the prompt. Prompt engineering is not simply a question-and-answer process; using clear, precise, and concise prompting formats significantly improves the quality of the output [25]. For instance, "Please find relevant information about Company A" is less clear and precise than "Please find the headquarters location, founder, main business, and founding year of Company A, and provide a 100-word company profile supported by references." The latter prompt yields results that are more aligned with the user's needs. In addition to this, prompt engineering also includes techniques such as one-shot or few-shot prompts, Chain of Thought (CoT), Reasoning and Acting (ReAct), and Tree of Thoughts (ToT) prompts [26].

Retrieval-Augmented Generation (RAG) [27] is a technique that leverages external knowledge bases to improve the accuracy of LLM outputs. It is one of the effective methods for addressing the issue of "hallucinations" in LLMs, especially when handling domain-specific or knowledge-intensive tasks. Currently, RAG is widely applied in knowledge graph construction, text summarization, and question-answering systems. RAG can be categorized into three types: Naive RAG, Advanced RAG, and Modular RAG.

Naive RAG, the first method to gain attention since the launch of ChatGPT, involves three steps: indexing, retrieval, and generation. Advanced RAG improves upon Naive RAG by adding pre-retrieval and post-retrieval strategies, addressing its limitations in retrieval precision, recall, hallucinations, and the issue of disjointed or incoherent output. Modular RAG builds on the foundations of the previous two approaches, offering superior adaptability and flexibility. Restructured RAG and rearranged RAG pipelines have been incorporated to tackle specific challenges, going beyond the fixed structures of Naive RAG and Advanced RAG.

Fine-tuning is the process of adjusting the parameters of a pre-trained large language model to a specific task or domain[20, 21]. When LLMs perform poorly on a specific task, it becomes necessary to consider fine-tuning the model, by fine-tuning a model on a small specific dataset, users can improve the LLMs' performance on this specific task. According to the OpenAI Platform[1] fine-tuning has at least four advantages, i.e., higher quality results than prompting, the ability to train on more examples than prompting, token saving, and lower latency requests. Some research has demonstrated the above mentioned advantages to an extent, for example, Schmirler R, et al. have found that task-specific supervised fine-tuning almost always improves downstream predictions, thus they suggest to always try fine-tuning, in particular for problems with small datasets[21]. Furthermore, there are many techniques and models for this work, such as full parameter, layer-specific, component-based, and multi-stage fine-tuning methods, as well as the LoRA, qLoRA techniques, models like GPT-4, GPT-3.5-turbo, T5, among other, are covered in reference [19], which provides a vast, high quality details on this information and the LLMs.

RAG, prompt engineering, and fine-tuning are commonly used methods to improve the accuracy of LLM outputs. However, users are often perplexed about which one to choose, and as a result, these techniques are frequently compared[28]. According to Yunfan et al., "prompt engineering leverages a model's inherent capabilities with minimal necessity for external knowledge and model adaptation. RAG can be likened to providing a model with a tailored textbook for information retrieval, ideal for precise information retrieval tasks. In contrast, fine-tuning is comparable to a student internalizing knowledge over time, suitable for scenarios requiring replication of specific structures, styles, or formats" [27], as shown in Fig.4. But, here is a tip from OpenAI on this issue that may help, i.e., try prompt engineering first, due to the lower investment of time and effort it requires.

---

[1] https://platform.openai.com/docs/guides/fine-tuning

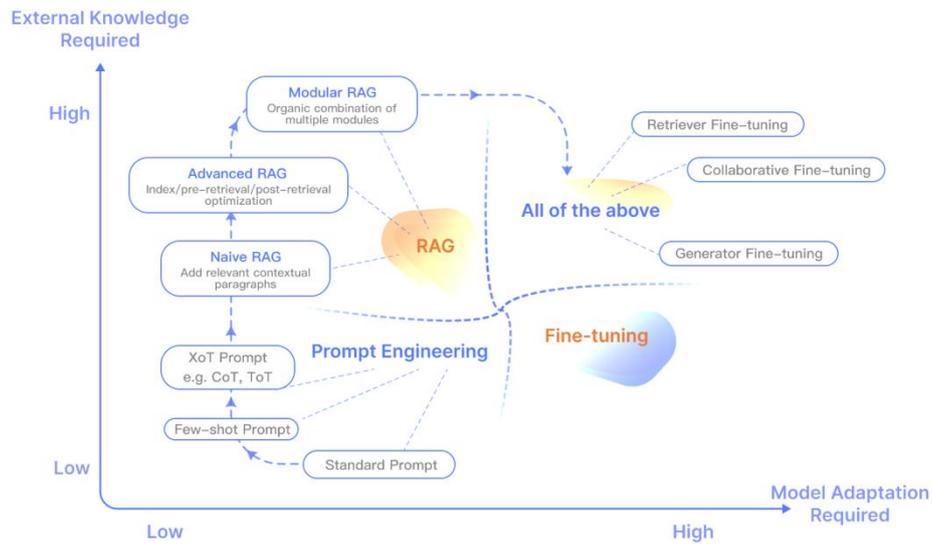

Figure 4. Technology tree of RAG research. *Source*: The figure is proposed from reference [27].

In the very first stage, the LLM is trained in a self-supervised manner on a large corpus to predict the next tokens given the input, which essentially means finding a good "initialization point" for the model parameters. This idea was originally widely used in the field of computer vision, where large-scale labeled image datasets, such as ImageNet, were used to initialize the parameters of vision models. To pre-train large language models, a vast amount of text data needs to be prepared, which must undergo rigorous cleaning to remove any potentially harmful or toxic content. After cleaning, the data is tokenized into a stream and split into batches for pre-training the language model [19]. Since the foundational capabilities of large language models mainly come from pre-training data, data collection and cleaning have a significant impact on the model's performance.

Tool learning refers to the process that aims to unleash the power of LLMs to effectively interact with various tools to accomplish complex tasks [29, 30]. Large-scale models do not inherently possess the ability to utilize APIs for forwarding generated text to designated email accounts. Moreover, since the data employed during the pre-training phase is not current but rather from a specific time frame, it is difficult for them to automatically retrieve up-to-date information from the web. Tool learning offers an effective solution to this limitation by seamlessly integrating large models with API interfaces. This integration allows for the execution of straightforward tasks such as automated email responses and real-time weather checks, as well as more complex tasks involving the reconstruction of workflows.

These technologies together lay the groundwork for LLMs, enabling them to perform impressively across a multitude of tasks and fields. With ongoing research, these technologies are continually being refined and enhanced to tackle the challenges that large models face in real-world applications.

## 2 A brief introduction of SciSci

SciSci, often referred to as "science of science", seeks to understand, quantify and predict scientific research and the resulting outcomes [31]. This includes analyzing the innovation process [32-34], measuring the influence of scientific publications [33, 35, 36], researchers [37, 38], journals [39, 40], and institutions [38, 41], as well as modeling the scientific collaboration an citation patterns [42, 43]. Additionally, it involves classifying various scientific domains [44, 45] and evaluating funding and success [46, 47]. The insights garnered from SciSci hold significant implications for management science and policy-making.

### 2.1 Typical SciSci studies

The fundamental concept underlying the development of SciSci is, notably, citation [48]. Citations serve as evidence by linking a researcher's work to demonstrate the validity of the authors' ideas. They create connections between authors, ideas, journals, institutions, and even countries, enabling the construction of citation networks or the application of citation counts for research evaluation purposes.

The introduction of the Science Citation Index (SCI) database in the 1950s significantly advanced citation analysis,with Price [49] among the early pioneers recognizing the importance of interconnected networks of scholarly papers. Although the SCI was initially intended to facilitate more effective literature searches for researchers, its immense potential in research evaluation soon became apparent. Phenomenon such as "cumulative advantage" [50], the "Matthew effect" [51], and "invisible colleges" [52] were observed and identified through citation analysis. Co-citation analysis [53], bibliographic coupling [54], and direct citation analysis [55] emerged, along with their derived forms, including author-level, journal-level, and keyword-level citation analysis.Regarding indicators, citation counts, h-index, journal impact factor, and their variants have been the most commonly utilized metrics for policy-making and research evaluation, despite ongoing criticisms.

Recently, metrics such as usage, tweets, and mentions, collectively referred to as "altmetrics," have been considered supplementary to traditional citations in research evaluation. Beyond their role in impact assessment, several research teams began focusing on knowledge mapping during the mid-1980s. Tools like Pajek and Ucinet were developed to facilitate the visualization of large networks. Boyack, Klavans, and Börner [56] were the first to map the backbone of science in 2005. More recent visualization tools specifically designed for SciSci, such as CiteSpace, VOSviewer, and CitNetExplorer, have made network generation more accessible to users.

**2.2 Recent advances in SciSci**

In the late 1990s, a group of computer scientists and physicists, with foresight, ventured into this field, new methods, new tools have been introduced into the field, the data source has also been greatly expanded. thus greatly broadening the disciplinary scope of SciSci. By 2017 and 2018, seminal works such as "The science of science: from the perspective of complex systems" [31], and "science of science" [57] popularized the term "science of science," attracting the attention of researchers from various disciplines, including physics, social sciences, mathematics, and information and computer science.

These newly engaged researchers approached the study of science as a complex system, consisting of numerous components and interactions. In this framework, components are represented by nodes, while interactions are depicted as links. Following the introduction of small-world and scale-free networks at the turn of the 20th century, interest surged in Graph Neural Networks (GNNs)[58, 59], multilayer networks [60], and hypergraph [61].

Graph Neural Networks (GNNs), alongside their variants including Graph Convolutional Networks (GCNs), Graph Attention Networks (GATs), and GraphSAGE, have demonstrated remarkable performance across a variety of SciSci tasks in recent years [62]. For instance, Huang, et al.,[63] highlighted the importance of paper classification in literature retrieval and bibliometric analysis, noting that traditional text-based approaches—such as those relying solely on keywords, titles, and abstracts—often overlook valuable information contained within cited papers. To address this gap, they introduced an improved GNN model aimed at enhancing the accuracy of paper classification. To tackle the issue that most citation dynamic models focus solely on individual nodes rather than the entire citation structure, Feng, et al.,[64] proposed a method to learn the entire information cascade process as the input of a sequential deep neural network.

Multilayer networks excel at capturing the complex relationships inherent in scientific activities, such as citation networks, collaboration networks, and institutional networks [65]. Science can be conceptualized as a complex system, comprising components with interactions. Traditional methods that represent these networks as single aggregated structures inevitably lead to information loss. To mitigate this issue, Wang, et al.,[66] combined co-citation networks, direct citation networks, and coupling networks into a multilayer network to predict potential academic collaborations in the field of gene editing. Their findings indicated that the multilayer network approach produced more accurate predictions than traditional collaboration network models.

Hypergraphs, which extend traditional graph structures, have gained recognition in the field of SciSci for their capacity to model complex, higher-order interactions. Contrary to Wang, et al.'s approach [66], some researchers [67] advocate for viewing academic collaboration through the lens of team dynamics rather than merely as interactions between pairs of agents. In this context, hypergraphs or bipartite graphs are seen as more insightful alternatives to traditional frameworks, which are limited to representing relationships between pairs. These researchers also promote an integrated approach that considers both semantic and structural features in academic collaboration. Such a holistic perspective is essential for achieving a comprehensive understanding of the intricate patterns and outcomes of scholarly interactions.

In summary, physicists and computer scientists have made significant contributions to the advancement of SciSci by applying domain-specific methods and tools and integrating them with established research topics. As a result, the scope of data in SciSci has evolved from abstract databases like Web of Science and PubMed to include platforms such as Mendeley, OpenAlex, and Overton. Figure 5 illustrates the common data types utilized in SciSci, providing insights into their nature and examples of sources from which they can be derived.

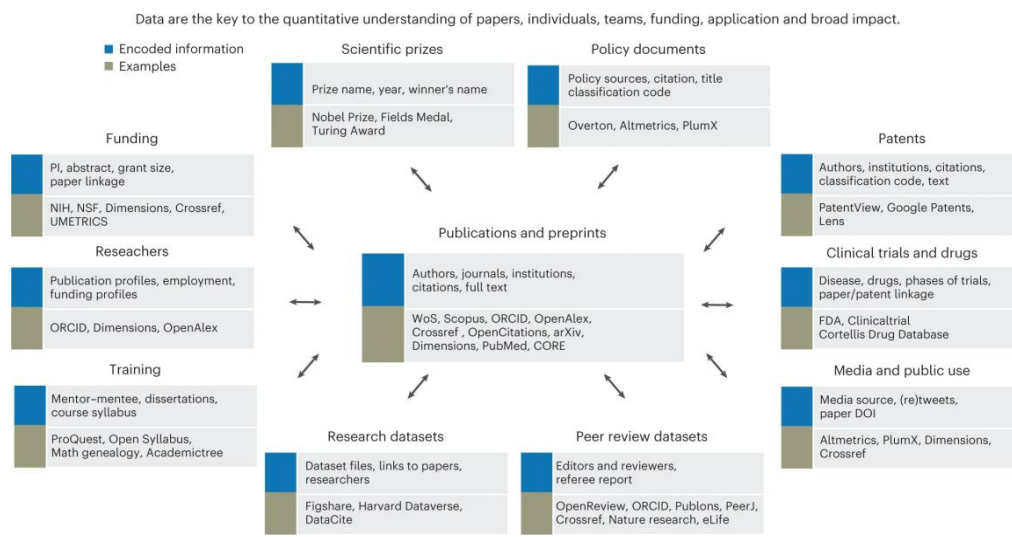

Figure 5. Commonly used data types in SciSci research. *Source*: The figure is proposed from reference [68].

## 3 The potential applications of LLMs in the field of SciSci

Since SciSci primarily focuses on the understanding, quantification, and prediction of science [31], this section discusses the impact of LLMs on SciSci from three perspectives: scientific perception, scientific evaluation, and scientific forecasting.

### 3.1 Scientific perception

In this context, scientific perception refers to the process through which individuals interpret and understand information derived from scientific literature. To enhance the comprehension of scientific phenomena, researchers have observed and statistically described the Matthew Effect in research productivity, and developed a suite of methods to map topics and semantic-enhanced themes [69, 70].

One traditional approach to knowledge topic extraction in SciSci involves generating co-word association maps based on frequently occurring words extracted from paper titles and keywords [69]. Essentially, co-word analysis involves extracting entities within sentences and establishing connections based on their relationships, which results in the formation of a single-mode network. One of the significant advantages of LLMs is their ability to efficiently extract entities and relationships from unstructured data, such as the full text of research papers in PDF format. This capability allows for a more comprehensive approach to data extraction compared to traditional methods. Once entities and their relationships are identified using LLMs, these can be visualized and manipulated within resulting networks.

In LLM-based entity relationship extraction, the relationships between entities are imbued with semantic dimensional information, presenting a richer and more nuanced array of information compared to networks constructed solely through traditional co-word methods. Additionally, the scope of entity relationship extraction can expand beyond just titles and keywords to encompass the entirety of research papers. This broadened scope significantly enhances the richness and complexity of the topics derived, thereby improving our understanding of the various dimensions of scientific knowledge. In summary, the integration of LLMs into SciSci offers new opportunities for deepening scientific perception by providing more sophisticated methods for topic extraction, relationship mapping, and data visualization, ultimately leading to a more comprehensive understanding of science as a complex system.

There are several ways to leverage LLMs to enhance traditional co-word analysis. For instance, techniques such as one-shot and few-shot prompting, along with prompt engineering in models like ChatGPT, can be employed to extract insights more effectively. Alternatively, users can directly call the API to access these capabilities [71]. To facilitate understanding, we have created a simple demonstration (see Fig.6). The code for this demo is freely available on GitHub[2].

This demo is built upon the Kimi LLM framework and illustrates the entire process of entity relationship extraction using prompts, along with the visualization of the results through the networkX library. It is important to note that this demo serves as a basic example to showcase the feasibility of extracting entity relationships from unstructured text using LLMs for knowledge graph construction. For those looking to improve the accuracy and effectiveness of their results, we recommend exploring additional features such as tool/function calling and the JSON Model [71, 72]. By refining these techniques, users can enhance the precision and utility of the knowledge graphs they create.

---

[2] https://github.com/Gqiang-Liang/Simple-demo-for-NRE/tree/main

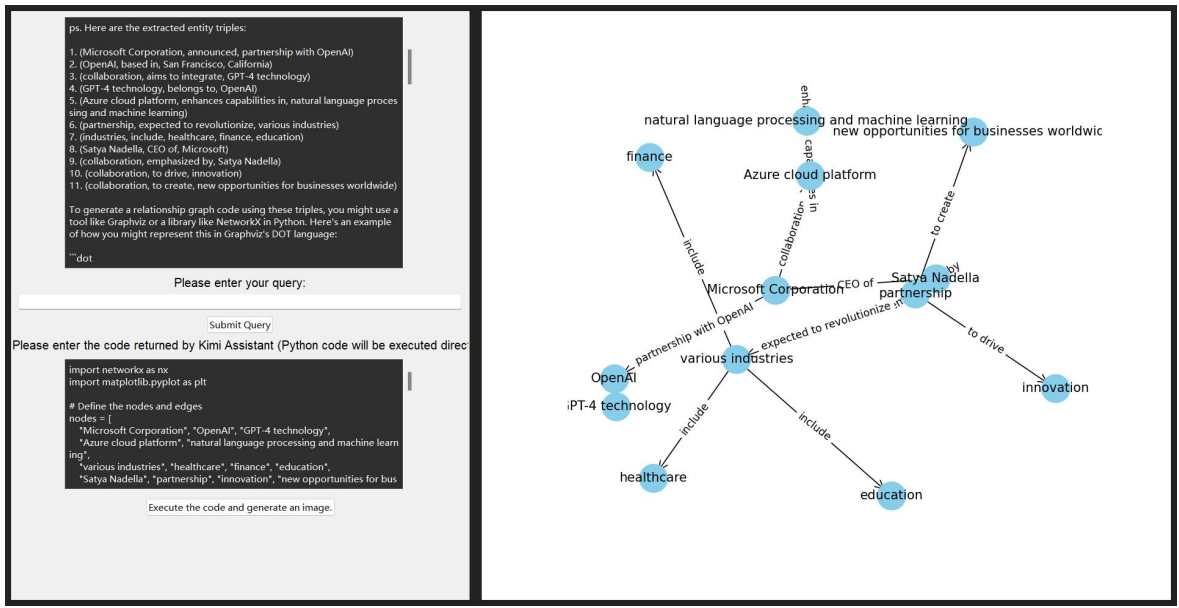

Figure 6. Demo utilizing Kimi for entity and relationship extraction, with networkX employed to visualize the results.

**3.2 Scientific evaluation**

The evaluation of research work by universities, institutions, journals, researchers, and individual research articles has become a routine aspect of modern society. These evaluations aim to enhance understanding of scientific activities, such as monitoring and managing performance, disseminating contributions, justifying public expenditures by demonstrating research value to taxpayers and stakeholders, and informing funding decisions [73].

Typical approaches for scientific evaluation include peer review and a range of quantitative methods, such as bibliometrics, complex networks analysis, and deep learning techniques. With the ongoing advancements in LLMs, we propose that the implementation of AI agents for scientific evaluation processes will emerge as a prominent direction in SciSci. To clarify the concept of AI agents, we represent them mathematically as follows: AI agents = LLMs + a set of skills (such as memory, function calling, and tool usage). The authors of this study have developed a "transformative research evaluation AI agent" based on ChatGLM during initial explorations[3]. However, it is recognized that the effectiveness of this AI agent still requires significant improvement. Nevertheless, these early explorations lay the groundwork for AI agent-based scientific evaluations in the field of SciSci.

By employing such AI agents, it becomes feasible to measure the influence of scientific publications, researchers, journals, and institutions. For instance, Fig.7 illustrates the interface of an AI agent developed on the Dify platform, showcasing its potential application in the evaluation landscape. As these AI agents continue to evolve, they promise to transform the metrics and methods used in evaluating scientific research and its impact across various domains.

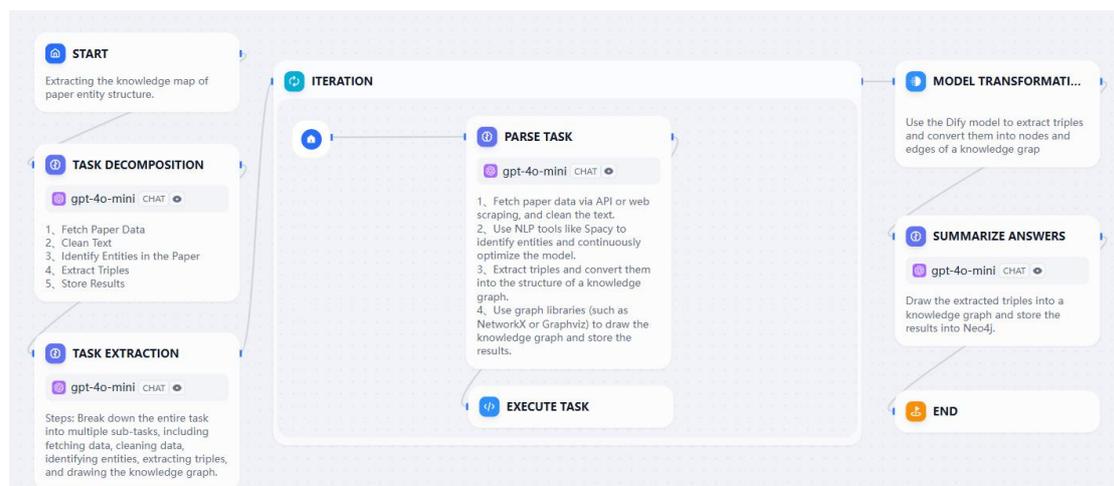

Figure 7. The designation for an AI agent based on the Dify platform

## 3.3 Scientific forecasting

Forecasting has always been at the forefront of planning and decision making, individuals and organisations seeking to maximise utilities and minimise risks. As trends and interests in scientific research evolve over time, it is vital to identify and forecast the trends and the future direction of development. Research communities have developed a series of tools to identify the trends and evolution of science, such as the iFORA system developed by National Research University Higher School of Economics [74], the Xinghuo Scientific Assistant[4] based on the Xinghuo LLM which powered by iFLYTEK Co.Ltd. In SciSci, the academic success of researchers remains an everlasting topic of significant importance in management science and policy-making [75]. In the future, the integration of LLMs into scientific research forecasting is expected to provide substantial opportunities for advancements in SciSci and to represent a significant transformation of traditional SciSci methodologies.

---

[3] https://chatglm.cn/main/gdetail/6632ecfeace21f9ff21cf4c0?lang=zh

Research fronts represent the cutting edge and growth frontier of scientific inquiry, having now become a focal point in global scientific and technological competition. Traditional forecasting methods employ co-citation clusters, co-citation clusters supplemented with citing articles, or direct citation clusters. Here, we propose an LLM-based multilayer network approach for forecasting research fronts. We evaluated current mainstream LLMs—including GPT-4o, Moonshoot-V1-8k, QwQ-32B-Preview, Gemini-Pro-1.5, and Deepseek-V3—and ultimately selected DeepSeek-V3 for multilayer network construction based on input/output costs, topic relevance, processing speed, and other key metrics (performance comparison shown in Table 2)

Table 2. Performance of mainstream LLMs

---

[4]https://paperlogin.iflytek.com/

| LLMS | Input cost ($/per million tokens) | Output cost ($/million tokens) | Output topic relativeness | Processing speed (second/paper) | Supported context length |
|---|---|---|---|---|---|
| GPT-4o | 2.5 | 10 | High | 8.5±2.1 | 128k |
| Deepseek-V3 | 0.14 | 0.28 | High | 10.8±3.5 | 64k |
| Gemini-Pro-1.5 | 3.5 | 10.5 | Low | 6.2±1.8 | 2M |
| Moonshoot-V1-8k | 1.66 | 1.66 | General | 12.4±4.2 | 8k |
| QwQ-32B-Preview | 0.12 | 0.18 | General | 15.6±5.0 | 8k |

We extracted Subject-Action-Object structures from publications using DeepSeek-V3 and constructed a multilayer network (figure 8) with the PyMnet toolkit, thereby facilitating research front forecasting.

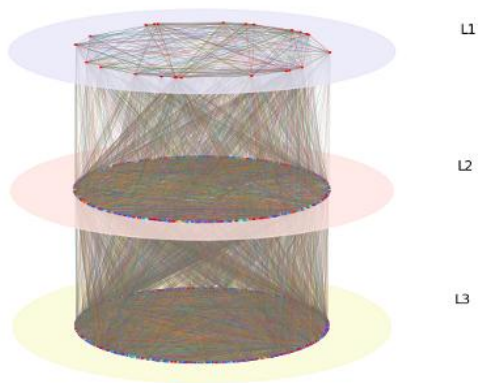

Figure 8. Research front forecasting by LLM-based multilayer network

**4 Conclusions**

The rapid advancement of LLMs presents significant opportunities for the evolution of SciSci. Researchers in this field can harness LLMs to explore previously unresolved research questions and to identify strategies for enhancing efficiency, particularly in areas such as name disambiguation. Furthermore, LLMs facilitate automated scientific research evaluation and trend prediction through the deployment of AI agents. However, these advancements also introduce challenges for traditional scientometricians. On one hand, the rise of LLMs calls for a deeper understanding and enhanced proficiency in computer technologies, including reinforcement learning and deep learning. On the other hand, it necessitates a reevaluation and redesign of existing theories and frameworks, potentially leading to the development of new tools and metrics in response to the AI era.

This paper offers a systematic review of the evolution of SciSci, the key technologies underlying LLMs, and the prospective applications of LLMs within this field. Due to space constraints, many potential applications of LLMs in specific domains of SciSci remain underexplored in this article. Examples include the integration of LLMs with full-text analysis, their combination with tasks such as sentiment analysis, semantic analysis, and text classification, their enhancement of citation analysis, and their potential to usher in an era of multimodal SciSci. These avenues are ripe for future exploration and can further enrich the landscape of SciSci research.